\documentclass[oribibl]{llncs}
\usepackage{multicol}
\usepackage{amsmath}
\usepackage{amssymb}
\usepackage{graphicx}
\usepackage{color}
\usepackage{hhline}
\usepackage{enumfrom}

\usepackage{hyperref}

\newcommand{\ignore}[1]{}

\newcommand{\msf}[1]{\mbox{\sf  #1}}

\newcommand{\resp}{\mbox{\sf Resp}}
\newcommand{\xresp}{\mbox{\sf x-Resp}}

\title{Reasoning about Counterfactuals and Explanations: Problems, Results and Directions }

\author{{\bf Leopoldo Bertossi}}
\institute{\vspace{-2mm}\bf Universidad Adolfo Ib\'a\~nez\\{\bf Faculty of Engineering and Sciences}\\ and\\ Millennium Inst. for Foundational Research on Data (IMFD)\\
 Santiago, \ Chile\\ leopoldo.bertossi@uai.cl}

\ignore{
\author{{\bf Leopoldo Bertossi}}
\institute{\bf Universidad Adolfo Ib\'a\~nez\\{\bf Faculty of Engineering and Sciences}\\ and\\ Millennium Inst. for Foundational Research on Data (IMFD)\\
 Santiago, \ Chile\\ leopoldo.bertossi@uai.cl}
 }

\begin{document}

\thispagestyle{empty}
\pagestyle{plain}
\maketitle

\begin{abstract} \vspace{-4mm}
 There are some recent approaches and results about the use of {\em answer-set programming} for specifying  counterfactual interventions on entities under classification, and reasoning about them. These approaches are flexible and modular in that they allow the seamless addition of domain knowledge. Reasoning is enabled by query answering from the answer-set program. The programs can be used to specify and compute {\em responsibility-based} numerical scores as attributive explanations for classification results.
\end{abstract}

\paragraph{\bf 1. Introduction.}
In this short paper we describe at a high level recent research that we have carried out in the area of score-based explanations to outcomes from classification models. We also describe how declarative specifications of- and reasoning with counterfactual explanations leading to score computation are enabled and supported by {\em answer-set programs} \cite{asp}.  References are given for the technical details. We also discuss some relevant research directions.

\paragraph{\bf 2. Attribution Scores.}
 Counterfactuals are at the very basis of the notion of \emph{actual causality} \cite{HP05}. They are hypothetical interventions (or changes) on variables of a causal structural model. Counterfactuals can be used to define and assign \emph{responsibility scores} to the variables in the model, with the purpose of quantifying the strength of their causal contribution  to a particular outcome \cite{CH04,halpern15}. These generals notions of actual causality have been applied in databases, to investigate actual causes and responsibilities for query results \cite{suciu,suciuDEBull,tocs}.

 Numerical scores have been applied in \emph{explainable AI}, and most prominently in machine learning models for classification \cite{molnar}. The general idea is that feature values in entities under classification are given numerical scores, to indicate how relevant they are for the outcome of the classification. For example, one might want to know how important is the city or the neighborhood where a client lives when a bank uses a classification algorithm to accept a loan request or not. This can be done by assigning a number to the feature value, e.g. to ``$\msf{Bronx in New York City}$". As such, it is a \emph{local explanation}, for the entity at hand, and in relation to all its participating feature values.

 A widely used score is $\msf{Shap}$ \cite{LetA20}, which is based on the Shapley value that is  used in coalition game theory \cite{R88}. It is based on \emph{implicit} counterfactuals and a numerical aggregation of the outcomes from the classification of those different counterfactual versions of the initial entity. Accordingly, the emphasis is not on the possible counterfactuals, but on the final numerical score. However, counterfactuals are interesting \emph{per se}. For example, we might want to know if the client, by changing his/her address, might turn a rejection into acceptance of the loan request. The so generated new entity, with a new address and a new label, is a \emph{counterfactual version} of the original entity.

 The $\xresp$ score was introduced in \cite{tplp}. It is defined in terms of explicit counterfactuals and responsibility as found in general actual causality. A more general version of it, the $\resp$ score, was introduced in \cite{deem}, and was compared with other scores, among them, $\msf{Shap}$.

\paragraph{\bf 3. Reasoning with Counterfactual and ASPs.}
 Taking seriously the idea that counterfactuals are  interesting  in their own right, {\em counterfactual intervention programs} (CIPs) were proposed in \cite{tplp}. They are \emph{answer-set programs} (ASPs) \cite{asp,gelfond} that specify counterfactual versions of an initial entity, reason about them, and compute $\xresp$ scores for feature values.

 Answer-set programming is a flexible and powerful logic programming para-digm that, as such, allows for declarative specifications and reasoning from them.  The (non-monotonic) semantics of a program is given in terms of its {\em stable models}, i.e. special models that make the program true \cite{GL91}. In our applications, the relevant counterfactual versions correspond to different models of the CIP.

 CIPs can be used to specify the relevant counterfactuals (by imposing extra conditions in rule bodies or using {\em program constraints} that filter models where they are violated), specify ``minimum-change" counterfactuals (by using {\em weak program constraints} that filter models where they are not minimally violated \cite{leone}), analyze different versions of them, and use them to specify and compute the $\xresp$ score. In particular,  one can specify and compute maximum-responsibility counterfactuals (through the use of weak program constraints \cite{tplp,rw21}.

 For our examples with decision-trees  \cite{tplp,rw21} and  with naive-Bayes classifiers \cite{ijclr21Ext}, we have used the {\em DLV} system (and its extensions) \cite{leone} that implements the ASP semantics.
 The classifiers can be specified directly inside the CIP, or can be invoked as external predicates \cite{tplp}. The latter case is useful when we interact with a {\em black-box classifier} \cite{rudin}, to which scores such as $\msf{Shap}$ and $\xresp$ can be applied.

 \paragraph{\bf 4. Semantics and Domain Knowledge.}
 CIPs are very flexible in that one can easily add \emph{domain knowledge} or \emph{domain semantics}, in such a way that certain counterfactuals are not considered, or others are privileged. With CIPs, many kinds of changes on the specification that are of potential interest can be easily and seamlessly  tried out on-the-fly, for exploration purposes \cite{ijclr21Ext,rw21}. All these changes and alternatives are much more difficult to implement with a purely procedural approach.

 In particular, one can specify domain-dependent {\em actionable counterfactuals} \cite{tplp}, that, in certain applications, make more sense or may lead to feasible changes of feature values for an entity to reverse a classification result \cite{ustun,karimi}.

 The definitions of attribution scores explicitly or implicity consider all counterfactual version of the entity under explanation. However, both their definition and computation should be influenced by the domain semantics, which could lead to ignore some counterfactuals or to give more importance to others. This could be done by declaratively specifying which is the case for different counterfactuals. Probabilistic constraints could be declared and imposed, affecting the underlying population, to which counterfactual versions belong (c.f. Section 6. below).

\paragraph{\bf 5. Queries and Reasoning.}
 Reasoning is enabled by query answering, for which two semantics are offered. Under the {\em brave semantics} one obtains as query answers those that hold in {\em some} model of the CIP. This can be useful to detect if there is minimum-change counterfactual version of the initial entity where the city is changed together with the salary.

 Under the {\em cautious semantics} one obtains answers that hold in all the models of the CIP, which could be used to identify feature values that have to be changed no matter what if we want to reverse the outcome.

 As components of a same program, we could, for example, interact at the same time with two different classifiers. It would be easy to compare their classifications and counterfactuals by means of query answering.

 Query answering on ASPs offers many opportunities. Actually, there have been some efforts to design and investigate query languages for explanations \cite{sube}. \ ASP offers a  query language for this task, and as a part of the same system that does the reasoning and computation \cite{ijclr21Ext}. The investigation of its full potential (or shortcomings) for this tasks remains to be carried out. This analysis has to done more on the basis of practical needs than at that of the expressive power of the query language (which has been investigated in the case of ASP \cite{dantsin}).

 \paragraph{\bf 6. Room for Probabilistic Reasoning.}
Attribution scores are usually of a probabilistic nature in that they consider a -possibly implicit- distribution on the entity population. This is the case of {\sf Shap}, {\sf Resp}; and also the {\sf Causal-Effect}, used in \cite{tapp16} for tuple-attribution w.r.t. query answering in databases. The distribution is an important element to consider when analyzing the complexity of score computation \cite{aaai21,guy,deem}.

 The first generation of answer-sets programming, the one that is mostly used, is not probabilistic, and does not provide much support for probabilistic reasoning. With some difficulty, one can do probabilistic reasoning through numerical aggregations (as with naive-Bayes classifiers in \cite{ijclr21Ext}).

 A probabilistic extension of a logic-based declarative semantics, as is the case of {\em ProbLog} \cite{luc},  would be welcome for ASP.
Actually, there are probabilistic extensions of the ASP-semantics \cite{baral} that could be tried in this direction, and not only for probabilistic classifiers, probabilistic counterfactual reasoning, or probabilistic score computation, but also for exploring semantic changes and conditions that are reflected on modified  distributions \cite{tplp}. The need for systems for probabilistic-ASP reasoning becomes crucial.


 \paragraph{\bf 7. Contexts and Interpretations.} When producing explanations, one should have in mind who is going to receive and  analyze them, in particular, those based on attribution scores. The final user, possibly a non-expert in explanation methodologies, has to make sense of them. For this reason, explanations should be conveyed {\em in terms of the context} of this user. Through this context the user will be in position to {\em interpret the explanations}. We have argued that formal ontologies are appropriate for describing and specifying  contexts \cite{mostafa,birte}. For this purpose, the ASP-based specification and computation of explanations  could interact ``at a similar logical level" with formal ontologies, e.g. conveying results from the former to the latter. This is a promising research direction. The integration of ASP and ontologies has been considered in  general terms (cf. \cite{mantas} for a discussion and references).

 \vspace{3mm} \noindent {\bf Acknowledgments: }  Part of this work was funded by ANID - Millennium Science Initiative Program - Code ICN17002.

\end{document}